\documentclass[letterpaper]{article} 
\usepackage{aaai25}  
\usepackage{times}  
\usepackage{helvet}  
\usepackage{courier}  
\usepackage[hyphens]{url}  
\usepackage{graphicx} 
\usepackage{natbib}  
\usepackage{caption} 
\usepackage{algorithm}
\usepackage{algorithmic}
\usepackage{amsmath}
\usepackage{multirow}
\usepackage{amssymb}

\newcommand\boldp[1]{\vspace{.1cm}\noindent\textbf{#1.}\hspace{.065cm}}
\usepackage{newfloat}
\usepackage{listings}
\DeclareCaptionStyle{ruled}{labelfont=normalfont,labelsep=colon,strut=off} 
\floatstyle{ruled}
\newfloat{listing}{tb}{lst}{}
\floatname{listing}{Listing}
\title{Spatial Annealing for Efficient Few-shot Neural Rendering}
\author{
    Yuru Xiao, Deming Zhai, Wenbo Zhao, Kui Jiang, Junjun Jiang, Xianming Liu\thanks{Corresponding authors}
}
\affiliations{
    Harbin Institute of Technology\\

    hitxyr@stu.hit.edu.cn, (zhaideming, wbzhao, jiangkui, jiangjunjun, csxm)@hit.edu.cn
}

\usepackage{bibentry}

\usepackage{subfigure}
\begin{document}

\maketitle

\begin{abstract}
Neural Radiance Fields (NeRF) with hybrid representations have shown impressive capabilities for novel view synthesis, delivering high efficiency. Nonetheless, their performance significantly drops with sparse input views. Various regularization strategies have been devised to address these challenges. However, these strategies either require additional rendering costs or involve complex pipeline designs, leading to a loss of training efficiency. Although FreeNeRF has introduced an efficient frequency annealing strategy, its operation on frequency positional encoding is incompatible with the efficient hybrid representations. In this paper, we introduce an accurate and efficient few-shot neural rendering method named \textbf{S}patial \textbf{A}nnealing regularized \textbf{NeRF} (\textbf{SANeRF}), which adopts the pre-filtering design of a hybrid representation. 
We initially establish the analytical formulation of the frequency band limit for a hybrid architecture by deducing its filtering process. Based on this analysis, we propose a universal form of frequency annealing in the spatial domain, which can be implemented by modulating the sampling kernel to exponentially shrink from an initial one with a narrow grid tangent kernel spectrum. This methodology is crucial for stabilizing the early stages of the training phase and significantly contributes to enhancing the subsequent process of detail refinement.
Our extensive experiments reveal that, by adding merely one line of code, SANeRF delivers superior rendering quality and much faster reconstruction speed compared to current few-shot neural rendering methods. Notably, SANeRF outperforms FreeNeRF on the Blender dataset, achieving 700$\times$ faster reconstruction speed. Code is available at https://github.com/pulangk97/SANeRF.
\end{abstract}

%
\section{Introduction}
\label{Introduction}

Neural Radiance Field (NeRF) \cite{mildenhall2021nerf}, represented by a multi-layer perception (MLP), is a powerful learning-based tool for detailed 3D reconstruction and high-fidelity novel view synthesis. However, NeRF requires dense input views or days of training time, limiting its application in real-world scenarios such as autonomous driving, robotics, and AR/VR. Although grid-based hybrid presentations \cite{muller2022instant, hu2023tri} have been introduced to improve training efficiency by accelerating convergence and reducing training time from hours to minutes, they still suffer from abnormal convergence and weak geometry awareness when dealing with sparse input views, a problem known as few-shot neural rendering.

Current few-shot neural rendering methods prioritize mitigating overfitting and enhancing geometry reconstruction, yet often neglect reconstruction efficiency. For instance, RegNeRF \cite{niemeyer2022regnerf}, DietNeRF \cite{jain2021putting}, and VGOS \cite{sun2023vgos} utilize patch-based geometry regularization or semantic consistency regularization, which involves computationally intensive forward processing on randomly sampled rendered patches. Alternatively, some methods require substantial modifications to the base architecture \cite{kim2022infonerf, zhu2024vanilla} or depend on 3D prior from pre-trained networks \cite{wang2023sparsenerf}, reducing pipeline efficiency. There is an urgent need for methods that integrate into existing efficient architectures to improve robustness against sparse views while maintaining computational efficiency.

FreeNeRF \cite{yang2023freenerf} adopts a coarse-to-fine training approach to address the few-shot neural rendering problem, akin to implicit geometry regularization. While FreeNeRF significantly enhances rendering performance in few-shot scenarios with minimal code adjustments, its training process remains time-consuming and labor-intensive due to its MLP-based architecture. Moreover, its implementation using frequency positional encoding is incompatible with hybrid acceleration architectures. Exploring an efficient way to embed the frequency annealing concept into hybrid-represented neural fields holds the promise of enhancing both practicality and performance.

Introducing frequency annealing in hybrid representations requires explicitly modeling the frequency band-limit of hybrid represented fields. While current methods employ pre-filtering strategies \cite{hu2023tri,liu2024rip} to mitigate high-frequency information for anti-aliasing, they often lack explicit analysis and formulation of the frequency bandwidth. Therefore, conducting frequency band limit analysis and modeling for grid-based neural fields is crucial for advancing the efficiency of few-shot neural rendering.

In this paper, we derive the explicit formation of frequency bandwidth for a pre-filtering-driven hybrid representation through an analysis of the filtering process. Building on this analysis, we introduce a more generalized form of frequency regularization in the spatial domain, termed spatial annealing, by examining the relationship between spatial area sampling and frequency bandwidth. Our method is applicable to both implicit and hybrid representations and can be implemented with just one line of code on two well-known pre-filtering-driven neural fields: MipNeRF and TriMipRF, characterized by implicit and hybrid representations, respectively. Extensive experiments on synthetic datasets and Blender \cite{mildenhall2021nerf} in a few-shot context demonstrate the effectiveness and efficiency of our approach. It not only outperforms the original TriMipRF by approximately 3dB in PSNR but also surpasses the state-of-the-art few-shot neural rendering technique, FreeNeRF \cite{yang2023freenerf}, by 0.3dB in PSNR. Additionally, our method achieves a training speed that is 700$\times$ faster than FreeNeRF, marking a significant advancement in both performance and efficiency. The main contributions of our work are summarized as follows:

\begin{itemize}
    \item We establish the analytical formulation of the frequency band limit for a pre-filtering designed hybrid representation architecture by analyzing its filtering process.

    \item We explore a general form of frequency annealing in the spatial domain based on frequency band limit analysis, which can be integrated into both implicit and hybrid representations with just one line of code. 

    \item We present extensive experimental results demonstrating that our method achieves superior efficiency and enhanced performance. Specifically, it outperforms FreeNeRF by 0.3 dB in PSNR on the Blender dataset while achieving training speeds 700 times faster.

\end{itemize}

\section{Related Work}
\label{relate}
\boldp{Neural Radiance Fields} 
Neural Radiance Field (NeRF) \cite{mildenhall2021nerf} is a distinguished 3D representation method renowned for its ability to render realistic novel views. It utilizes a Multilayer Perceptron (MLP) network to implicitly associate 3D coordinates with radiance attributes, including density and color. Over the past few years, the research community has developed numerous NeRF extensions, enhancing its application across a variety of domains such as novel view synthesis \cite{martin2021nerf, barron2021mip, barron2022mip, mildenhall2022nerf}, surface reconstruction \cite{oechsle2021unisurf, wang2021neus, wang2023pet}, dynamic scene modeling \cite{fang2022fast, liu2023robust, park2021nerfies}, and 3D content generation \cite{gu2023nerfdiff, poole2022dreamfusion, shue20233d, chen2023single, hong2023lrm}. Although NeRF has pioneered several advancements across different research areas, it encounters significant challenges. The method particularly struggles with slow reconstruction speed and a heavy dependence on the density of input views, limiting its efficiency and practicality in broader applications. These challenges highlight the need for ongoing research and development to optimize NeRF's performance and expand its usability in the field of 3D representation and beyond.

\boldp{Few-shot Neural Rendering} 
NeRF grapples with accurately fitting the low-frequency geometry when processing sparse input views. Many approaches incorporate 3D information like sparse point clouds \cite{deng2022depth}, estimated depth \cite{wang2023sparsenerf, cao2022fwd, wei2021nerfingmvs, roessle2022dense}, or dense correspondences \cite{truong2023sparf, lao2024corresnerf} for enhanced supervision. However, integrating additional algorithms or models to acquire extra 3D data adds complexity and implementation challenges to the overall process. Beyond leveraging 3D information, some methods aim to directly regularize geometry using strategies such as random patch-based semantic consistency \cite{jain2021putting}, or geometric regularization \cite{kwak2023geconerf, niemeyer2022regnerf, sun2023vgos, seo2023mixnerf}. These techniques, however, tend to prolong training times because of the added rendering burden on extra sampled patches. Learning-based methods \cite{lin2023vision, yu2021pixelnerf, chen2021mvsnerf, wang2021ibrnet, long2022sparseneus, liu2022neural, chibane2021stereo, trevithick2021grf}, on the other hand, seek to train a network on extensive multi-view datasets to internalize a geometric prior, yet these approaches require costly pre-training and additional fine-tuning steps.

Recently, FreeNeRF \cite{yang2023freenerf} addresses the issue of under-constrained geometry in sparse view settings by progressively increasing the frequency of positional encoding. This technique eliminates the need for additional 3D information and does not increase the computational cost of the base architecture, positioning it as a robust baseline for few-shot neural rendering applications. While the coarse-to-fine training approach of FreeNeRF has demonstrated effectiveness in few-shot scenarios, it still necessitates lengthy training periods. Furthermore, its method of applying a mask to frequency positional encoding is not readily adaptable to various NeRF acceleration architectures. Motivated by FreeNeRF's achievements, our goal is to develop a straightforward solution for a NeRF acceleration architecture, aiming to enhance performance in few-shot settings without compromising training efficiency.

\section{Preliminaries and Motivation}
\boldp{Frequency Regularization}
FreeNeRF \cite{yang2023freenerf} addresses the few-shot challenge by progressively increasing the frequency of positional encoding throughout the training process, a technique known as frequency regularization. This approach is straightforwardly implemented using a modulated mask as
\begin{equation}
    \gamma^\prime(t, T, \mathbf{x}) = \gamma(\mathbf{x}) \odot \mathbf{M}(t, T, L)
    \label{eq:ec}
\end{equation}
with
\begin{equation}
    {M}_i(t, T, L) = \begin{cases}1 & \text{if } i \le \frac{t \cdot L}{T} + 3 \\ \frac{t \cdot L}{T} - \lfloor\frac{t \cdot L}{T}\rfloor & \text{if } \frac{t \cdot L}{T} + 3 < i \le \frac{t \cdot L}{T} + 6 \\ 0 & \text{if } i > \frac{t \cdot L}{T} + 6\end{cases},
    \label{eq:mask}
\end{equation}
where $t$ and $T$ represent the current and total iteration steps, respectively, $\gamma$ and $\gamma^\prime$ correspond to the initial and masked positional encodings, $L$ controls the maximum frequency of positional encoding, and ${M}_i$ is the $i$-th element of mask $\mathbf{M}$, which linearly expands throughout the training process.

\boldp{TriMipRF} TriMipRF \cite{hu2023tri} introduces a pre-filtering strategy for tri-plane representation. Due to the incompatibility of integrated positional encoding (IPE) introduced by MipNeRF \cite{barron2021mip} with this hybrid model, TriMipRF performs area-sampling by querying features on the three planes with plane levels correlating with the projected radius of the sampling sphere within the cone. The relevant levels are denoted by $\lfloor l \rfloor$ and $\lceil l \rceil$, where $l$ signifies the query level corresponding to the sphere's radius:
\begin{equation}
    \label{level0}
    l=\log_2{(\frac{\mathbf{\tau}}{\ddot{r}})},
\end{equation}
where $\ddot{r} = {T_0}/{\sqrt{\pi}}$ represents the base radius associated with base planes with maximum resolution $r_0$, $\mathbf{\tau}$ denotes the radius of the sampling sphere, and $T_0$ is the period of the base planes. The plane level $[l] \in \mathbb{N}$ has a relationship with the plane resolution as $r_{[l]} = r_0 / 2^{[l]}$. For detailed formation, please refer to TriMipRF \cite{hu2023tri}.

\section{Method}
\begin{figure*}[ht]
    \centering
    \includegraphics[width=\linewidth]{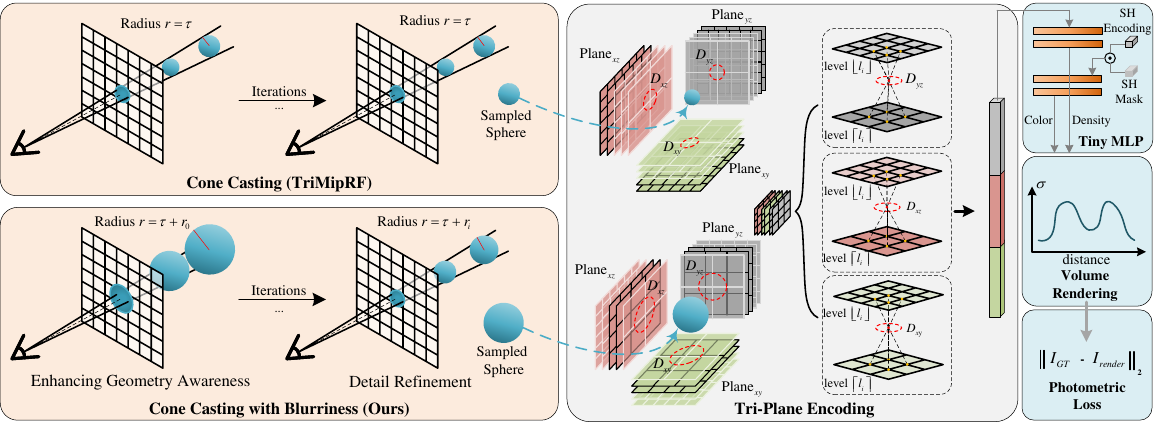}
    \caption{An overview of the complete framework. We introduce an efficient few-shot neural rendering method utilizing TriMipRF. Initially, we set the sample sphere's radius larger than that of the base sphere to optimize low-frequency geometry, as depicted in the bottom left corner. During training, we progressively reduce the sphere's radius through exponential decay, thereby refining local details within the reconstructed global structure.}
    \label{fig:method}
\end{figure*}
\subsection{Spatial Annealing in Implicit Representation}
\label{sas}
We start by examining the relationship between frequency regularization (see Eq. \ref{eq:ec}) and the pre-filtering strategy based on integrated positional encoding (IPE). We consider a multivariate Gaussian distribution in 3D space, depicted as
\begin{equation}
    G(\mathbf{x}) = e^{-\frac{1}{2}(\mathbf{x}-\boldsymbol{{\mu}})^{T} \boldsymbol{\Sigma}^{-1}  (\mathbf{x}-\boldsymbol{{\mu}})},
\end{equation}
where $\boldsymbol{\mu}$ represents the position, and $\boldsymbol{\Sigma}$ denotes the covariance matrix. For straightforward comparison, we model the Gaussian as isotropic, characterized by a diagonal $\boldsymbol{\Sigma}$ with uniform elements $\sigma_f^2$, reflecting the sample space size. Following MipNeRF \cite{barron2021mip}, we calculate the Gaussian's integrated positional encoding $\gamma_G$ as
\begin{equation}
    \gamma_G(\boldsymbol{\mu}) = \gamma'_G(\boldsymbol{\mu}) \odot \mathbf{M}_L
\end{equation}
with
\begin{equation}
    \label{imask}
    \begin{split}
        \mathbf{M}_L &= [e^{-\frac{1}{2}\sigma_f^2},e^{-\frac{1}{2}\sigma_f^2},...,e^{-2^{2L-3}\sigma_f^2},e^{-2^{2L-3}\sigma_f^2}] \\
        \gamma'_G(\boldsymbol{\mu}) &= [\sin(\boldsymbol{\mu}), \cos(\boldsymbol{\mu}), ..., \sin(2^{L-1}\boldsymbol{\mu}), \cos(2^{L-1}\boldsymbol{\mu})],
    \end{split}
\end{equation}
where $\gamma'_G(\boldsymbol{\mu})$ represents the frequency positional encoding of $\boldsymbol{\mu}$, and $\mathbf{M}_L$ denotes a low-pass mask applied to this encoding, with $L$ controlling the maximum encoded frequency. The mask's structure aligns with a Gaussian distribution, featuring a covariance $\sigma_i^2 = 1/\sigma_f^2$, where $\sigma_i^2$ regulates the frequency bandwidth.

Upon comparing Eq. \ref{imask} with Eq. \ref{eq:ec}, we observe a notable similarity between the IPE and frequency regularization: both apply a low-pass mask to the original positional encoding, with the mask's form being the primary difference. Drawing inspiration from FreeNeRF \cite{yang2023freenerf}, which linearly expands the frequency mask (effectively exponentially increasing the frequency bandwidth), we adopt an exponential growth model for $\sigma_i^2$, defined as $\sigma_i^2 = f_s2^{x}$, where $x$ is the step increment corresponding to the iteration count and $f_s$ controls the initial frequency bandwidth. Concurrently, the spatial Gaussian's covariance $\sigma_f^2$ decreases exponentially, represented as
\begin{equation}
    \label{de}
    \sigma_f^2 = \frac{1}{f_s2^x}.
\end{equation}
Consequently, we deduce that the frequency regularization introduced by FreeNeRF \cite{yang2023freenerf} can be executed by inversely adjusting the spatial sample space within the pre-filtering strategy, as detailed in Eq. \ref{de}. This insight guides the exploration of the form of spatial annealing strategy in hybrid representation. 

\subsection{Spatial Annealing in Hybrid Representation}
\boldp{Filtering Process} 
Deducing the relationship between the spatial sampling strategy and frequency bandwidth in hybrid representation requires a detailed examination of the filtering process. Building on recent research into grid-based models \cite{zhao2024grounding, shabanov2024banf}, we construct the mathematical form of the tri-plane represented field and deduce the filtering process of the base architecture, which subsequently motivates the establishment of the spatial annealing strategy in hybrid representation. 

Based on recent research, the band limit of an optimized field can be determined by the sampling kernel under specific conditions \cite{shabanov2024banf}. We analyze the sampling process of this optimized neural field directly. General grid-based models \cite{zhao2024grounding, shabanov2024banf} reconstruct the optimized continues field $\mathbf{f}(\mathbf{x})$ from discrete feature points by applying a convolution process with a kernel function $\kappa$:

\begin{equation}
    \mathbf{\hat{f}}(\mathbf{x}) = \hat{\kappa}_\omega(\mathbf{x}) * \sum_{n} \mathbf{f}(\mathbf{x}) \cdot \delta(\mathbf{x} - \mathbf{x}_n),
\end{equation}
where $\delta$ is the impulse function, $\omega$ is the frequency bandwidth of the kernel, and $x_n \in \mathbb{R}^3$ are discrete grid points. In current hybrid representation fields, the kernel function can be implemented using simple linear interpolation, equivalent to a conical kernel $\Lambda_{\omega}(\mathbf{x})$. For regular grids with period $T$, the frequency bandwidth of the kernel is approximated by $\omega = \frac{2\pi}{T}$, as the conical kernel is not an ideal low-pass filter.

In TriMipRF, the reconstructed signal $\mathbf{\hat{f}}^{m}(\mathbf{x},[l])$ for each plane $m$ at a specific level $[l]$ can be represented as
\begin{equation}
    \label{fxlm}
    \mathbf{\hat{f}}^m(\mathbf{x},[l]) =  \Lambda^m_{\omega_{[l]}}(\mathbf{x}) *  \sum_{n} \mathbf{f}^m(\mathbf{x}, [l]) \cdot \delta(\mathbf{x} - \mathbf{x}_{n}^{[l]}),
\end{equation}
where $m \in \{xy, xz, yz\}$ represents the label of each plane, $\Lambda^m_{\omega_{[l]}}(\mathbf{x})$ represents a two-dimensional conical kernel, and $\{\mathbf{x}_{n}^{[l]}\}$ is the uniform lattice with level $[l]$. The bandwidth of each plane at a specific level $[l]$ can be represented as $\omega_{[l]} = \frac{2\pi}{2^{[l]}T_0}$, where $T_0 = \frac{B_{max} - B_{min}}{r_0}$ represents the period of the grid with maximum resolution $r_0$ and bounding box size $B_{max} - B_{min}$. We label the bandwidth $\omega$ corresponding to the level $l$ as $\omega_l$. TriMipRF utilizes multi-resolution tri-planes to model varying levels of detail (LOD), it queries features from adjacent integral levels of $l$ corresponding to the size of the sphere (refer to Eq. \ref{level0}) as
\begin{equation}
    \label{eq:filterlevel}
    \mathbf{\Tilde{f}}^m(\mathbf{x}, l) = \Lambda_{\omega_c}(l) * \sum_{[l]} \mathbf{\hat{f}}^m(\mathbf{x},l) \cdot \delta(l - [l]),
\end{equation}
where $\Lambda_{\omega_c}(l)$ is a conical kernel with constant bandwidth $\omega_c$, and $[\cdot]$ represents discrete integral values. TriMipRF utilizes a pre-filtering strategy across the level dimension to filter out fields with higher-frequency bandwidths that cannot be well-sampled by discrete rays, thereby addressing aliasing artifacts. This approach constructs a continuous field along the level dimension within a continuous range of sampling spaces. Equation \ref{eq:filterlevel} can also be rewritten in the form of linear interpolation across the level dimension:
\begin{equation}
    \label{eq:comb}
    \mathbf{\Tilde{f}}^m(\mathbf{x},l) = \frac{\lceil l \rceil - l}{\lceil l \rceil - \lfloor l \rfloor} \mathbf{\hat{f}}^m(\mathbf{x},\lfloor l \rfloor) + \frac{l - \lfloor l \rfloor}{\lceil l \rceil - \lfloor l \rfloor} \mathbf{\hat{f}}^m(\mathbf{x},\lceil l \rceil).
\end{equation}
The continuous field represented by each feature plane can be considered a linear combination of the two fields at specific levels $\lceil l \rceil$ and $\lfloor l \rfloor$. Therefore, the bandwidth $\omega_m$ of the combined field lies within the range $(\omega_{\lceil l \rceil}, \omega_{\lfloor l \rfloor})$ as
\begin{equation}
    \label{eq:inequality}
    \omega_{\lceil l \rceil} \leq \omega_m \leq \omega_{\lfloor l \rfloor}.   
\end{equation}
The final feature field can be represented as the concatenation of components corresponding to each plane. Since the sampling process operates independently based on the same kernel and identical querying level $l$, the bandwidth $\omega$ of the final field is determined by $\omega_m$.

\boldp{Spatial Annealing Strategy}
We aim to explore the relationship between the frequency bandwidth $\omega_m$ and the size of spatial sampling space. Since the bandwidth of the field cannot be analytically represented, we approximate $\omega_m$ by
\begin{equation}
    \label{eq:app}
    \omega_m \approx \frac{2\pi}{2^{l}T_0}.
\end{equation}
The approximation consistently satisfies the inequality given by Eq. \ref{eq:inequality}, since Eq. \ref{eq:app} is monotonic along level $l$. Such an approximation has an error bound of $\omega_{\lfloor l \rfloor} - \omega_{\lceil l \rceil}$. Inspired by FreeNeRF, we exponentially increase the frequency bandwidth from an initial value $f_s$ as $\omega_m = f_s \cdot 2^x$, where $x$ is the step increment corresponding to the iteration count. As TriMipRF aligns level $l$ with the size of the sampling sphere according to Eq. \ref{level0}, the relationship between the size of the sphere $\tau$ and the increment step $x$ can be computed as
\begin{equation}
    \label{eq:decrehybrid}
    \tau = \frac{2\sqrt{\pi}}{f_s2^x}.
\end{equation}
We observe that Eq. \ref{eq:decrehybrid} and Eq. \ref{de} exhibit similar forms, both exponentially decreasing the size of the sampling space from an initial value. Consequently, we establish a universal form of frequency regularization in the spatial domain that is suitable for both implicit and hybrid representations.

\begin{figure}[!tp]
    \centering
    \includegraphics[width=1\linewidth]{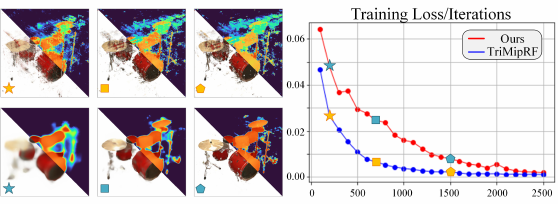}
    \caption{{Comparison results during the training procedure.} The training loss curve reveals that TriMipRF exhibits premature convergence early, resulting in the underfitting of the geometry depicted on the left. Conversely, our spatial annealing strategy effectively addresses this challenge.}
    \label{fig:c2f}
\end{figure}
Fig. \ref{fig:method} provides an overview of our method. We have developed a coarse-to-fine training strategy based on TriMipRF \cite{hu2023tri}. Initially, we implement blurring in the rendering process by increasing the radius $r$ of the sample sphere, as depicted in the bottom-left corner of Fig. \ref{fig:method}. At the beginning of training, this approach aids in optimizing the reconstruction of global geometric structures, crucial for mitigating overfitting issues common in few-shot scenarios, as illustrated in Fig. \ref{fig:c2f}. The radius of the sample area is systematically reduced following an exponential decay described in Eq. \ref{eq:decrehybrid}. This gradual reduction aims to progressively shift the training focus towards enhancing local geometric and textural details. The specific steps of this annealing process are outlined as
\begin{equation}
    r_i=\begin{cases}\tau+\dfrac{f_s}{2^{\upsilon x}},i<T\\\tau,i\geq T\end{cases},x=\left\lfloor\dfrac{iN_{split}}{T}\right\rfloor,
\end{equation}
where $N_{split}$ denotes the total number of decrement steps, $i$ represents the current iteration count, and $T$ indicates the stopping point of annealing. $f_s$ determines the initial size of the sphere, while $\vartheta$ governs the rate of decrement.

\begin{figure}[!t]
    \centering
    \subfigure[GTK spatial]{
        \includegraphics[width=0.437\linewidth]{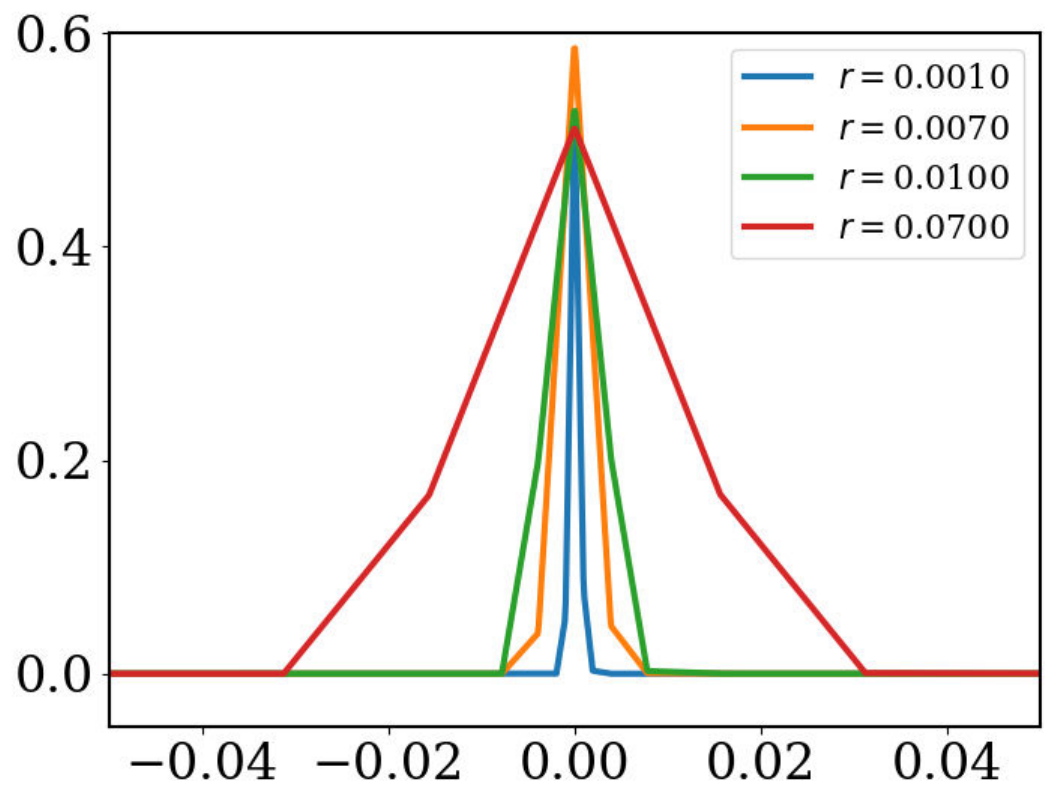}
        \label{fig:gtk_spatial}
    }
    \subfigure[GTK spectrum]{
        \includegraphics[width=0.45\linewidth]{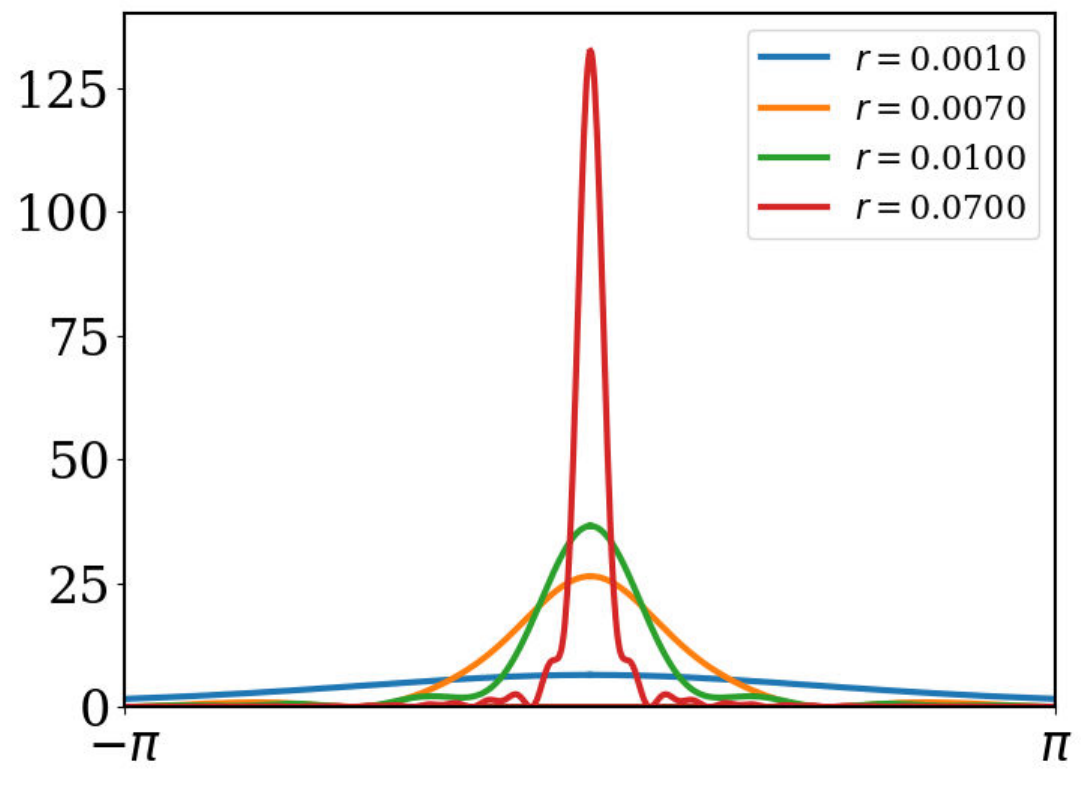}
        \label{fig:gtk_spectrum}
    }
    \caption{Visualization of GTK and its spectrum. We vary the size $r$ of the sampling sphere and measure the 1-D grid tangent kernel along with its frequency spectrum. The GTK spectrum indicates that a larger $r$, with a larger sampling space, corresponds to a narrower spectrum.}
    \label{fig:gtk}
\end{figure}
\boldp{Assessing the Geometry Awareness of Spatial Blurriness}
In this paper, we strategically modulate the size of the sample sphere starting from an initially large value. This approach introduces blurriness corresponding to low-frequency band limit (refer to Eq. \ref{eq:app}), which enhances geometry awareness during the initial training stage. To verify this insight, we conduct a spectrum analysis of the grid-based model utilizing grid tangent kernel (GTK) theory \cite{zhao2024grounding}. We initially simplified Eq. \ref{eq:comb} as follows:
\begin{align}
        \mathbf{\Tilde{f}}^m(\mathbf{x},l) = &\frac{\lceil l \rceil - l}{\lceil l \rceil - \lfloor l \rfloor} \sum_{n} \varphi^{\lfloor l \rfloor}(\mathbf{x}_{n}^{\lfloor l \rfloor})\boldsymbol{w}^{\lfloor l \rfloor}_n + \\
        &\frac{l - \lfloor l \rfloor}{\lceil l \rceil - \lfloor l \rfloor} \sum_{n} \varphi^{\lceil l \rceil}(\mathbf{x}_{n}^{\lceil l \rceil})\boldsymbol{w}^{\lceil l \rceil}_n , \nonumber
\end{align}

where $\boldsymbol{w}^{\lfloor l \rfloor}_n$ and $\boldsymbol{w}^{\lceil l \rceil}_n$ are features stored in grid points at levels $\lfloor l \rfloor$ and $\lceil l \rceil$, respectively, and $\varphi$ represents the interpolation weights. As the GTK of a grid-based model remains stationary during training \cite{zhao2024grounding}, the GTK during training can be computed by the optimized neural field:
\begin{equation}
    [\boldsymbol{G}_g(t)]_{i,j}=\left\langle\frac{\partial \mathbf{\Tilde{f}}^m(\mathbf{x}_i, l)}{\partial\boldsymbol{w}},\frac{\partial \mathbf{\Tilde{f}}^m(\mathbf{x}_j,l)}{\partial\boldsymbol{w}}\right\rangle , 
\end{equation}
where $\boldsymbol{w}$ denotes features stored in grid points. We numerically analyze the 1D spatial GTK of the TriMip-represented field and its spectrum, as shown in Fig. \ref{fig:gtk}. The results indicate that with a larger sampling sphere, the spectrum becomes narrower, which biases training more towards low-frequency information, especially low-frequency geometry \cite{zhao2024grounding}. Such spectral bias performs exceptionally well when handling extreme scenarios, such as sparse views that specifically lack multi-view geometry consistency. As the training iteration progresses, the size of the sample sphere exponentially decreases, corresponding to a wider GTK spectrum, which shifts the training focus towards high-frequency details. Although current research has reached this insight for implicit representation \cite{yang2023freenerf}, we expand this insight to grid-based hybrid representation utilizing recently introduced GTK analysis.

\begin{figure}[!t]
    \centering
    \includegraphics[width=1\linewidth]{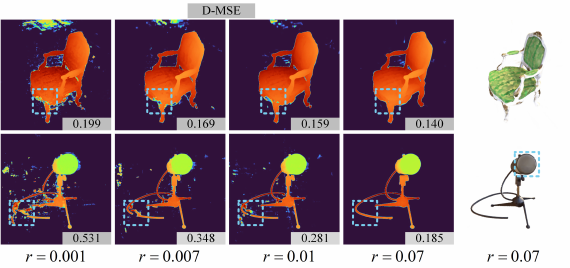}
    \caption{Qualitative rendered depth results with varying size of sampling sphere.}
    \label{fig:blurfs}
\end{figure}

We conduct experiments to validate this insight. We vary the sampling sphere size $r$ from a small value (0.001) to a relatively larger value (0.07), aligning with the setting of the GTK analysis, and evaluate the depth mean square error (D-MSE) between the rendered depth results and the ground truth (GT) depth maps, as shown in Fig. \ref{fig:blurfs}. We observe that the D-MSE decreases with increasing sampling sphere size. Additionally, qualitative results indicate that a larger sampling area with blurred rendered results corresponds to more complete geometry.

\section{Experiments}
\label{experiments}
\boldp{Datasets and Metrics}
We evaluate our method using the Blender dataset \cite{mildenhall2021nerf}, which comprises 8 synthetic scenes observed from a surround view perspective. Consistent with the FreeNeRF \cite{yang2023freenerf} and DietNeRF \cite{jain2021putting}, we train our model on 8 views identified by the IDs 26, 86, 2, 55, 75, 93, 16, and 73. The evaluation is conducted on 25 images, evenly selected from the test set. All images are downsampled by a factor of 2 using an average filter. For quantitative analysis, we report the average scores across all test scenes for PSNR, SSIM, and LPIPS.

\boldp{Degree Truncation of Spherical Harmonic Encoding}
When the input views are sparse, fitting the view-dependent colors becomes challenging. To address this and cover a broader range of unseen view directions, we directly mask the higher levels of the Spherical Harmonic Encoding \cite{verbin2022ref}, which can be represented as
\begin{equation}
    {M}_i(n) =
    \left\{
        \begin{array}{l}
            1, i \leq \sum_{j=0}^{n-1} 2j+1 \\
            0, i > \sum_{j=0}^{n-1} 2j+1
        \end{array},
    \right.
\end{equation}
where $n$ represents the truncated level and ${M}_i$ is the $i$-th element of the Spherical Harmonic mask $\mathbf{M}$. The mask zeroes out all Spherical Harmonic components above level $n$ while retaining all lower-level components.

\boldp{Implementation Details}
We implement our methodology based on the TriMipRF codebase \cite{hu2023tri}, utilizing the PyTorch framework \cite{paszke2019pytorch}. Building on TriMipRF's original configurations, we introduce additional hyperparameters tailored to our spatial annealing strategy. Specifically, we initialize the sphere size $f_s$ at 0.15, set the decrease rate $\vartheta$ to 0.2, define the total number of decrement steps $N_{split}$ as 30, and establish the stop point $T$ at $2K$. In the few-shot setting, there is a significant reduction in the number of input rays. Consequently, we limit the maximum training steps for both our method and the baseline TriMipRF to one-tenth of those employed in TriMipRF's full-view setting. We train our model using the AdamW optimizer \cite{loshchilov2017decoupled} for $2.5K$ iterations, with a learning rate of $2 \times 10^{-3}$ and a weight decay of $1 \times 10^{-5}$. Regarding the degree truncation in Spherical Harmonic Encoding, we consistently set the truncated level $n$ to 2 across all experiments.

\begin{figure}[!tp]
    \centering
    \includegraphics[width=\linewidth]{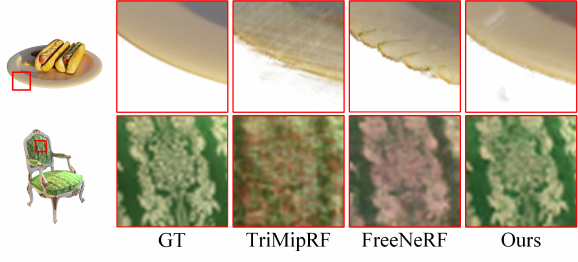}
    \caption{Qualitative results on Blender. We present qualitative comparisons of our method with the base architecture TriMipRF and the FreeNeRF few-shot baseline, utilizing 8 input views consistent with the FreeNeRF configuration.}
    \label{fig:blender1}
\end{figure}

\subsection{Blender Dataset}
Fig. \ref{fig:blender1} and Tab. \ref{tab:blender1} show the qualitative and quantitative comparisons on the Blender dataset \cite{mildenhall2021nerf}, respectively. The qualitative analysis reveals that the TriMipRF \cite{hu2023tri} produces noticeable distorted rendered results in the ``chair" and ``hotdog" scenes. In contrast, our approach, which requires only a few additional lines of code, effectively addresses these issues. When compared with the latest few-shot baseline, FreeNeRF \cite{yang2023freenerf}, our method shows superior performance, particularly in detail-rich areas highlighted in red boxes. Specific examples include the texture in the ``chair" scene and the edge of the plate in the ``hotdog" scene.

Our method demonstrates significant advancements in quantitative results, outperforming TriMipRF \cite{hu2023tri} by nearly 3 dB in PSNR with similar 35 seconds reconstruction time. 
Furthermore, the evaluated training durations demonstrate that our method achieves faster reconstruction speed. Notably, while the FreeNeRF \cite{yang2023freenerf} necessitates 7.5 hours for training, and DietNeRF \cite{jain2021putting}, with its patch-based semantic regularizer, requires an extensive 26 hours, our approach achieves a remarkable 700$\times$ reconstruction speed compared to FreeNeRF. In addition to its efficiency, our method also delivers superior quantitative outcomes, exceeding FreeNeRF by 0.27 dB in PSNR. While these results are impressive, extending the training duration has the potential to further enhance performance. To this end, we increase the training iterations to $5K$, while keeping all other settings unchanged. As shown in Tab. \ref{tab:blender1}, our method not only surpasses FreeNeRF with a 0.55 dB gain in PSNR but also delivers a substantial 350$\times$ acceleration.

\begin{table}[!tp]
    \centering
    \small
    \setlength{\tabcolsep}{1mm} 
    {
    \begin{tabular}{l||ccccr}
    
    \hline
    Method & PSNR$\uparrow$ & SSIM$\uparrow$ & LPIPS$\downarrow$ & Train$\downarrow$   \\
    \hline
    NeRF    & 14.93 & 0.687 & 0.318 & {6 h} \\
    MipNeRF     & 18.74 & 0.830 &0.240 & {9 h} \\
    Simplified NeRF     & 20.09 & 0.822 & 0.179 & {2 h} \\


    TriMipRF      & 21.63 & 0.818 & 0.183  & \textbf{35 s}\\

    \hline
    
    DietNeRF    & 23.15 & 0.866 & 0.109 & 26 h\\
    DietNeRF, $\mathcal{L}_\mathbf{MSE}$ ft    & 23.59 & 0.874 & {0.097} & 32 h\\
    InfoNeRF    & 22.01 & 0.852 & 0.133 & 7 h \\
    MixNeRF   & 23.84 & 0.878 & 0.103 & 9 h \\
    FreeNeRF   & 24.26 & {0.883} &   {0.098}   & 7.5 h  \\

    \hline
    VGOS   & 21.32 & 0.861 & 0.168 & 240 s \\
    FSGS   & 24.64 & \textbf{0.895} & {0.095} & 180 s \\
    DNGaussian   & 24.31 & 0.886 & \textbf{0.088} & 110 s \\
    Ours ($2.5K$ iterations)      & 24.53 & 0.881 & 0.102 & \textbf{35 s}\\
    Ours ($5K$ iterations)     & \textbf{24.81} & 0.882 & 0.101 & 70 s\\
    \hline
    \end{tabular}}
    \caption{Quantitative results on Blender with 8 input views.}
    \label{tab:blender1}
\end{table}

\section{Method Analysis}

\subsection{Frequency Regularization vs Spatial Annealing}
\label{fre&spa}

In previous section, we provide a theoretical analysis that clarifies the relationship between the pre-filtering strategy and frequency regularization based on integrated positional encoding. To demonstrate the strategy's validity and to evaluate our method's performance across various base architectures, we apply it using the JAX MipNeRF framework \cite{barron2021mip}. In this setup, we adjust the initial sphere size to 0.2, establish the decrease rate $\vartheta$ at 1, and set the total number of decrease steps $N_{split}$ to 33. 

\begin{figure}[!htp]
    \centering
    \includegraphics[width=\linewidth]{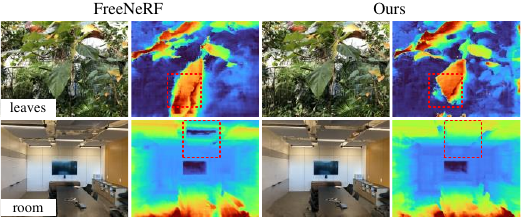}
    \caption{Qualitative comparisons on LLFF with 3 input views. We implement our method utilizing the JAX MipNeRF architecture to thoroughly evaluate our method's effectiveness across different base architectures.}
    \label{fig:llff_mip}
\end{figure}



    

\begin{table}[!htp]
    \centering

    \small
    \setlength{\tabcolsep}{1mm} 
    {
    \begin{tabular}{l|ccc|ccc}  \hline
    \multirow{2}{*}{Method} &  \multicolumn{3}{c|}{LLFF} & \multicolumn{3}{c}{DTU}\\ 
    &  PSNR$\uparrow$ & SSIM$\uparrow$  & LPIPS$\downarrow$ & PSNR$\uparrow$ & SSIM$\uparrow$ & LPIPS$\downarrow$ \\ \hline

    Mip-NeRF  &  14.62          & 0.351          & 0.495   & 7.64 & 0.227 & 0.655 \\ 

    RegNeRF  & 19.08     & 0.587          & 0.336     & 15.33 & 0.621 & 0.341 \\ 
    FreeNeRF &  {19.63}          &  {0.612}          & 0.308  & 18.02 & \textbf{0.680} & \textbf{0.318}\\ 

    Ours  &   \textbf{19.67} &  \textbf{0.616} & \textbf{0.288}   &  \textbf{18.15} & 0.675 & 0.320  \\ \hline
    
    \end{tabular}}
    \caption{Quantitative results on LLFF and DTU with 3 input views. All baseline methods, along with our method, are implemented using the JAX MipNeRF architecture.}
    \label{tab:LLFF}
\end{table}

Fig. \ref{fig:llff_mip} and Tab. \ref{tab:LLFF} present qualitative and quantitative comparisons of our method with FreeNeRF \cite{yang2023freenerf}. Quantitatively, our method matches or slightly outperforms FreeNeRF, showing up to a 0.05 dB and 0.1 dB increase in PSNR on the LLFF dataset \cite{mildenhall2019local} and the DTU dataset \cite{jensen2014large}, respectively. Qualitatively, our method provides improved novel views and depth maps. For instance, whereas FreeNeRF struggles with accurately reconstructing the leaf edges in the first row and the ceiling's middle section in the second row, our method maintains geometric consistency and offers more detailed depth maps. This advantage likely stems from our method's greater adaptability. Unlike frequency regularization, constrained by a fixed frequency range, our method dynamically adjusts the sample space to a wider range, thereby enhancing global geometry reconstruction.

\subsection{Robustness to View Sparsity}
\label{robust}

\begin{figure}[!htp]
    \centering

    \subfigure[PSNR]{
        \includegraphics[width=0.45\linewidth]{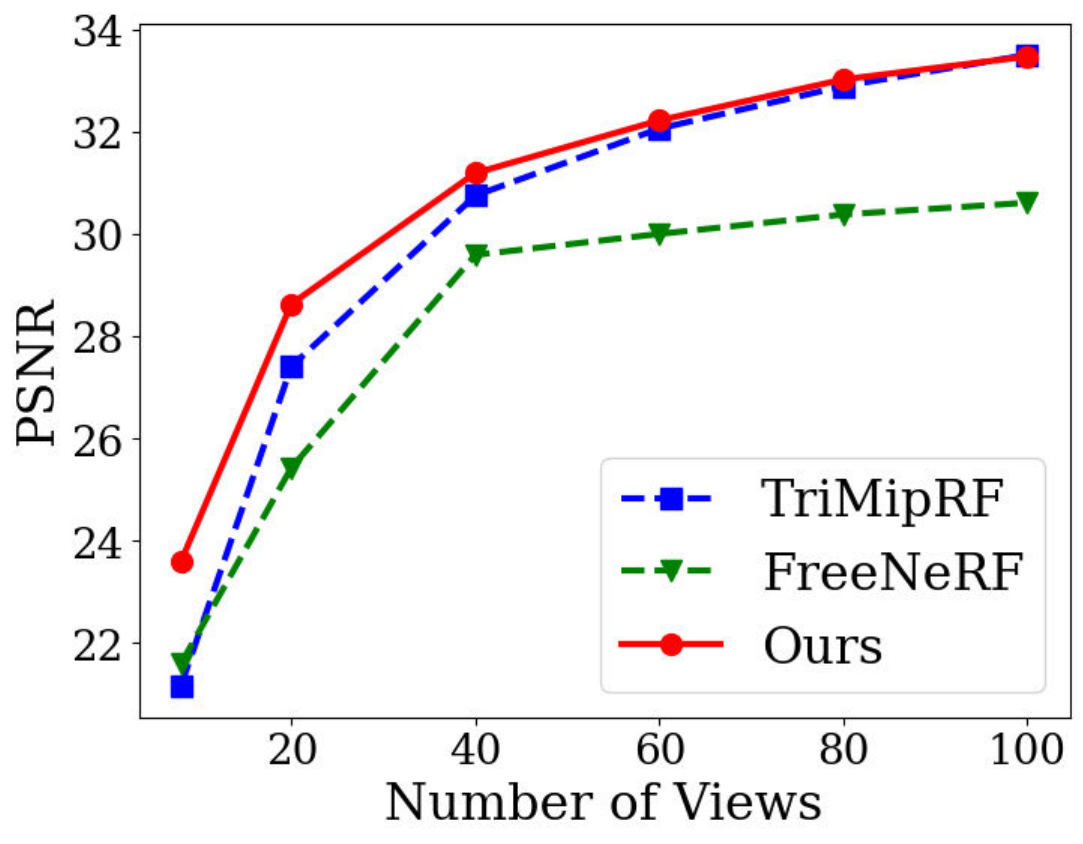}
        \label{fig:psnr_robust}
    }\subfigure[SSIM]{
        \includegraphics[width=0.48\linewidth]{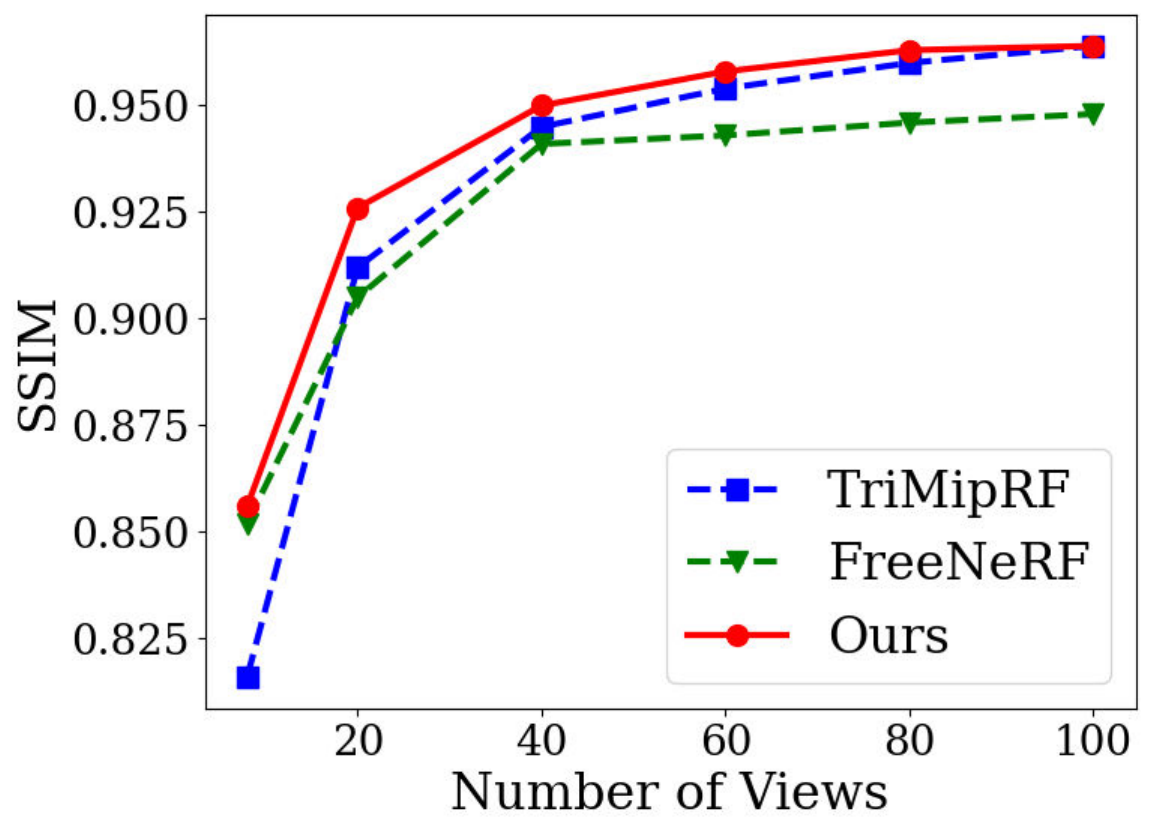}
        \label{fig:ssim_robust}
    }
    \caption{Validation of SANeRF's robustness to view sparsity compared with TriMipRF and FreeNeRF.}
    \label{fig:robust}
\end{figure}

We validate our method alongside the TriMipRF \cite{hu2023tri} using varying numbers of input views to assess our method's resilience to view sparsity, as shown in Fig. \ref{fig:robust}. We use the first $n$ images from the Blender's training set as input, where $n$ is varied among 8, 20, 40, 60, 80, and 100, with the dataset containing a total of 100 images. We fix the total iteration steps at 10,000 and the spatial annealing strategy's endpoint at 2,000. The curve of Fig. \ref{fig:robust} demonstrates that our method outperforms TriMipRF under different input view conditions. Notably, it shows marked improvements in scenarios with sparse input views, achieving a 2.50 dB and 1.22 dB higher PSNR with 8 and 20 input views, respectively. 

\subsection{Ablation Studies}
\label{ablation}

\begin{figure}[!tp]
    \centering
    \includegraphics[width=\linewidth]{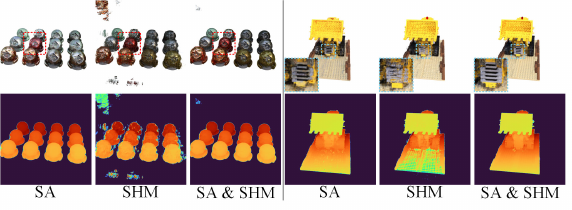}
    \caption{Qualitative ablation study on the Blender dataset. Our proposed spatial annealing strategy (SA) demonstrates impressive rendered depth results, while the spherical harmonic mask (SHM) addresses color distortion.}
    \label{fig:abl}
\end{figure}

To thoroughly assess the efficacy of our proposed method, we conduct qualitative and quantitative ablation studies using the Blender dataset \cite{mildenhall2021nerf} with 8 input views, as detailed in Fig. \ref{fig:abl} and Tab. \ref{tab:abl}. We evaluate the contributions of the newly introduced spatial annealing strategy (SA) and the spherical harmonic mask (SHM) separately. The findings reveal that the spatial annealing strategy significantly enhances performance in a few-shot setting, achieving a 1.2 dB increase in PSNR on the Blender dataset with SA alone. Regarding qualitative results, the rendered RGB images and depth maps demonstrate that our spatial annealing strategy primarily enhances the reconstruction of low-frequency geometry but does not effectively address color distortions. Conversely, the SHM effectively mitigates color distortions but fails to resolve geometric underfitting. The combination of both strategies results in well-established geometry and photorealistic rendering. These results underscore the substantial improvements achieved by our method.

\begin{table}[!htp]
    \centering 

    \small
    \setlength{\tabcolsep}{1mm} 
    {
    \begin{tabular}{c|cc|ccc|c} 
    \hline

    & SA & SHM & PSNR$\uparrow$ & SSIM$\uparrow$ & LPIPS$\downarrow$  & Train\\ \hline
TriMipRF & $\times$  & $\times$  & 21.63 & 0.818  & 0.183 & 35 s\\ \hline
\multirow{3}{*}{Ours} & $\checkmark$  & $\times$  & 22.83 & 0.856 & 0.120 & \multirow{3}{*}{35 s} \\
                      & $\times$  & $\checkmark$  & 22.27 & 0.825 & 0.173 \\
                      & $\checkmark$  & $\checkmark$  & 24.53 & 0.881 & 0.102 \\ \hline
    \end{tabular}}
    \caption{Quantitative ablation study on the Blender dataset.}
    \label{tab:abl}
\end{table}

\section{Conclusion}
In this study, we present a novel spatial annealing strategy tailored for a hybrid architecture equipped with a pre-filtering strategy. This approach adaptively modifies the spatial sampling size using a meticulously designed annealing process, enhancing geometric reconstruction and detail refinement. Remarkably, our approach requires minimal modifications to the base architecture's code to achieve state-of-the-art performance in few-shot scenarios while preserving efficiency. We acknowledge that our spatial annealing strategy possesses the potential to enhance the training stability of diverse architectures crafted with pre-filtering.




\bibliography{aaai25}

\begin{thebibliography}{49}
\providecommand{\natexlab}[1]{#1}

\bibitem[{Barron et~al.(2021)Barron, Mildenhall, Tancik, Hedman, Martin-Brualla, and Srinivasan}]{barron2021mip}
Barron, J.~T.; Mildenhall, B.; Tancik, M.; Hedman, P.; Martin-Brualla, R.; and Srinivasan, P.~P. 2021.
\newblock Mip-nerf: A multiscale representation for anti-aliasing neural radiance fields.
\newblock In \emph{Proceedings of the IEEE/CVF International Conference on Computer Vision}, 5855--5864.

\bibitem[{Barron et~al.(2022)Barron, Mildenhall, Verbin, Srinivasan, and Hedman}]{barron2022mip}
Barron, J.~T.; Mildenhall, B.; Verbin, D.; Srinivasan, P.~P.; and Hedman, P. 2022.
\newblock Mip-nerf 360: Unbounded anti-aliased neural radiance fields.
\newblock In \emph{Proceedings of the IEEE/CVF Conference on Computer Vision and Pattern Recognition}, 5470--5479.

\bibitem[{Cao, Rockwell, and Johnson(2022)}]{cao2022fwd}
Cao, A.; Rockwell, C.; and Johnson, J. 2022.
\newblock Fwd: Real-time novel view synthesis with forward warping and depth.
\newblock In \emph{Proceedings of the IEEE/CVF Conference on Computer Vision and Pattern Recognition}, 15713--15724.

\bibitem[{Chen et~al.(2021)Chen, Xu, Zhao, Zhang, Xiang, Yu, and Su}]{chen2021mvsnerf}
Chen, A.; Xu, Z.; Zhao, F.; Zhang, X.; Xiang, F.; Yu, J.; and Su, H. 2021.
\newblock Mvsnerf: Fast generalizable radiance field reconstruction from multi-view stereo.
\newblock In \emph{Proceedings of the IEEE/CVF international conference on computer vision}, 14124--14133.

\bibitem[{Chen et~al.(2023)Chen, Gu, Chen, Tian, Tu, Liu, and Su}]{chen2023single}
Chen, H.; Gu, J.; Chen, A.; Tian, W.; Tu, Z.; Liu, L.; and Su, H. 2023.
\newblock Single-stage diffusion nerf: A unified approach to 3d generation and reconstruction.
\newblock In \emph{Proceedings of the IEEE/CVF International Conference on Computer Vision}, 2416--2425.

\bibitem[{Chibane et~al.(2021)Chibane, Bansal, Lazova, and Pons-Moll}]{chibane2021stereo}
Chibane, J.; Bansal, A.; Lazova, V.; and Pons-Moll, G. 2021.
\newblock Stereo radiance fields (srf): Learning view synthesis for sparse views of novel scenes.
\newblock In \emph{Proceedings of the IEEE/CVF Conference on Computer Vision and Pattern Recognition}, 7911--7920.

\bibitem[{Deng et~al.(2022)Deng, Liu, Zhu, and Ramanan}]{deng2022depth}
Deng, K.; Liu, A.; Zhu, J.-Y.; and Ramanan, D. 2022.
\newblock Depth-supervised nerf: Fewer views and faster training for free.
\newblock In \emph{Proceedings of the IEEE/CVF Conference on Computer Vision and Pattern Recognition}, 12882--12891.

\bibitem[{Fang et~al.(2022)Fang, Yi, Wang, Xie, Zhang, Liu, Nie{\ss}ner, and Tian}]{fang2022fast}
Fang, J.; Yi, T.; Wang, X.; Xie, L.; Zhang, X.; Liu, W.; Nie{\ss}ner, M.; and Tian, Q. 2022.
\newblock Fast dynamic radiance fields with time-aware neural voxels.
\newblock In \emph{SIGGRAPH Asia 2022 Conference Papers}, 1--9.

\bibitem[{Gu et~al.(2023)Gu, Trevithick, Lin, Susskind, Theobalt, Liu, and Ramamoorthi}]{gu2023nerfdiff}
Gu, J.; Trevithick, A.; Lin, K.-E.; Susskind, J.~M.; Theobalt, C.; Liu, L.; and Ramamoorthi, R. 2023.
\newblock Nerfdiff: Single-image view synthesis with nerf-guided distillation from 3d-aware diffusion.
\newblock In \emph{International Conference on Machine Learning}, 11808--11826. PMLR.

\bibitem[{Hong et~al.(2023)Hong, Zhang, Gu, Bi, Zhou, Liu, Liu, Sunkavalli, Bui, and Tan}]{hong2023lrm}
Hong, Y.; Zhang, K.; Gu, J.; Bi, S.; Zhou, Y.; Liu, D.; Liu, F.; Sunkavalli, K.; Bui, T.; and Tan, H. 2023.
\newblock Lrm: Large reconstruction model for single image to 3d.
\newblock \emph{arXiv preprint arXiv:2311.04400}.

\bibitem[{Hu et~al.(2023)Hu, Wang, Ma, Yang, Gao, Liu, and Ma}]{hu2023tri}
Hu, W.; Wang, Y.; Ma, L.; Yang, B.; Gao, L.; Liu, X.; and Ma, Y. 2023.
\newblock Tri-miprf: Tri-mip representation for efficient anti-aliasing neural radiance fields.
\newblock In \emph{Proceedings of the IEEE/CVF International Conference on Computer Vision}, 19774--19783.

\bibitem[{Jain, Tancik, and Abbeel(2021)}]{jain2021putting}
Jain, A.; Tancik, M.; and Abbeel, P. 2021.
\newblock Putting nerf on a diet: Semantically consistent few-shot view synthesis.
\newblock In \emph{Proceedings of the IEEE/CVF International Conference on Computer Vision}, 5885--5894.

\bibitem[{Jensen et~al.(2014)Jensen, Dahl, Vogiatzis, Tola, and Aan{\ae}s}]{jensen2014large}
Jensen, R.; Dahl, A.; Vogiatzis, G.; Tola, E.; and Aan{\ae}s, H. 2014.
\newblock Large scale multi-view stereopsis evaluation.
\newblock In \emph{Proceedings of the IEEE conference on computer vision and pattern recognition}, 406--413.

\bibitem[{Kim, Seo, and Han(2022)}]{kim2022infonerf}
Kim, M.; Seo, S.; and Han, B. 2022.
\newblock Infonerf: Ray entropy minimization for few-shot neural volume rendering.
\newblock In \emph{Proceedings of the IEEE/CVF Conference on Computer Vision and Pattern Recognition}, 12912--12921.

\bibitem[{Kwak, Song, and Kim(2023)}]{kwak2023geconerf}
Kwak, M.-S.; Song, J.; and Kim, S. 2023.
\newblock GeCoNeRF: few-shot neural radiance fields via geometric consistency.
\newblock In \emph{Proceedings of the 40th International Conference on Machine Learning}, 18023--18036.

\bibitem[{Lao et~al.(2024)Lao, Xu, Liu, Zhao et~al.}]{lao2024corresnerf}
Lao, Y.; Xu, X.; Liu, X.; Zhao, H.; et~al. 2024.
\newblock CorresNeRF: Image Correspondence Priors for Neural Radiance Fields.
\newblock \emph{Advances in Neural Information Processing Systems}, 36.

\bibitem[{Lin et~al.(2023)Lin, Lin, Lai, Lin, Shih, and Ramamoorthi}]{lin2023vision}
Lin, K.-E.; Lin, Y.-C.; Lai, W.-S.; Lin, T.-Y.; Shih, Y.-C.; and Ramamoorthi, R. 2023.
\newblock Vision transformer for nerf-based view synthesis from a single input image.
\newblock In \emph{Proceedings of the IEEE/CVF Winter Conference on Applications of Computer Vision}, 806--815.

\bibitem[{Liu et~al.(2024)Liu, Hu, Yang, Chen, Wang, Chen, Cai, Gao, and Zhao}]{liu2024rip}
Liu, J.; Hu, W.; Yang, Z.; Chen, J.; Wang, G.; Chen, X.; Cai, Y.; Gao, H.-a.; and Zhao, H. 2024.
\newblock Rip-NeRF: Anti-aliasing Radiance Fields with Ripmap-Encoded Platonic Solids.
\newblock \emph{arXiv preprint arXiv:2405.02386}.

\bibitem[{Liu et~al.(2022)Liu, Peng, Liu, Wang, Wang, Theobalt, Zhou, and Wang}]{liu2022neural}
Liu, Y.; Peng, S.; Liu, L.; Wang, Q.; Wang, P.; Theobalt, C.; Zhou, X.; and Wang, W. 2022.
\newblock Neural rays for occlusion-aware image-based rendering.
\newblock In \emph{Proceedings of the IEEE/CVF Conference on Computer Vision and Pattern Recognition}, 7824--7833.

\bibitem[{Liu et~al.(2023)Liu, Gao, Meuleman, Tseng, Saraf, Kim, Chuang, Kopf, and Huang}]{liu2023robust}
Liu, Y.-L.; Gao, C.; Meuleman, A.; Tseng, H.-Y.; Saraf, A.; Kim, C.; Chuang, Y.-Y.; Kopf, J.; and Huang, J.-B. 2023.
\newblock Robust dynamic radiance fields.
\newblock In \emph{Proceedings of the IEEE/CVF Conference on Computer Vision and Pattern Recognition}, 13--23.

\bibitem[{Long et~al.(2022)Long, Lin, Wang, Komura, and Wang}]{long2022sparseneus}
Long, X.; Lin, C.; Wang, P.; Komura, T.; and Wang, W. 2022.
\newblock Sparseneus: Fast generalizable neural surface reconstruction from sparse views.
\newblock In \emph{European Conference on Computer Vision}, 210--227. Springer.

\bibitem[{Loshchilov and Hutter(2017)}]{loshchilov2017decoupled}
Loshchilov, I.; and Hutter, F. 2017.
\newblock Decoupled weight decay regularization.
\newblock \emph{arXiv preprint arXiv:1711.05101}.

\bibitem[{Martin-Brualla et~al.(2021)Martin-Brualla, Radwan, Sajjadi, Barron, Dosovitskiy, and Duckworth}]{martin2021nerf}
Martin-Brualla, R.; Radwan, N.; Sajjadi, M.~S.; Barron, J.~T.; Dosovitskiy, A.; and Duckworth, D. 2021.
\newblock Nerf in the wild: Neural radiance fields for unconstrained photo collections.
\newblock In \emph{Proceedings of the IEEE/CVF Conference on Computer Vision and Pattern Recognition}, 7210--7219.

\bibitem[{Mildenhall et~al.(2022)Mildenhall, Hedman, Martin-Brualla, Srinivasan, and Barron}]{mildenhall2022nerf}
Mildenhall, B.; Hedman, P.; Martin-Brualla, R.; Srinivasan, P.~P.; and Barron, J.~T. 2022.
\newblock Nerf in the dark: High dynamic range view synthesis from noisy raw images.
\newblock In \emph{Proceedings of the IEEE/CVF Conference on Computer Vision and Pattern Recognition}, 16190--16199.

\bibitem[{Mildenhall et~al.(2019)Mildenhall, Srinivasan, Ortiz-Cayon, Kalantari, Ramamoorthi, Ng, and Kar}]{mildenhall2019local}
Mildenhall, B.; Srinivasan, P.~P.; Ortiz-Cayon, R.; Kalantari, N.~K.; Ramamoorthi, R.; Ng, R.; and Kar, A. 2019.
\newblock Local light field fusion: Practical view synthesis with prescriptive sampling guidelines.
\newblock \emph{ACM Transactions on Graphics (TOG)}, 38(4): 1--14.

\bibitem[{Mildenhall et~al.(2021)Mildenhall, Srinivasan, Tancik, Barron, Ramamoorthi, and Ng}]{mildenhall2021nerf}
Mildenhall, B.; Srinivasan, P.~P.; Tancik, M.; Barron, J.~T.; Ramamoorthi, R.; and Ng, R. 2021.
\newblock Nerf: Representing scenes as neural radiance fields for view synthesis.
\newblock \emph{Communications of the ACM}, 65(1): 99--106.

\bibitem[{M{\"u}ller et~al.(2022)M{\"u}ller, Evans, Schied, and Keller}]{muller2022instant}
M{\"u}ller, T.; Evans, A.; Schied, C.; and Keller, A. 2022.
\newblock Instant neural graphics primitives with a multiresolution hash encoding.
\newblock \emph{ACM transactions on graphics (TOG)}, 41(4): 1--15.

\bibitem[{Niemeyer et~al.(2022)Niemeyer, Barron, Mildenhall, Sajjadi, Geiger, and Radwan}]{niemeyer2022regnerf}
Niemeyer, M.; Barron, J.~T.; Mildenhall, B.; Sajjadi, M.~S.; Geiger, A.; and Radwan, N. 2022.
\newblock Regnerf: Regularizing neural radiance fields for view synthesis from sparse inputs.
\newblock In \emph{Proceedings of the IEEE/CVF Conference on Computer Vision and Pattern Recognition}, 5480--5490.

\bibitem[{Oechsle, Peng, and Geiger(2021)}]{oechsle2021unisurf}
Oechsle, M.; Peng, S.; and Geiger, A. 2021.
\newblock Unisurf: Unifying neural implicit surfaces and radiance fields for multi-view reconstruction.
\newblock In \emph{Proceedings of the IEEE/CVF International Conference on Computer Vision}, 5589--5599.

\bibitem[{Park et~al.(2021)Park, Sinha, Barron, Bouaziz, Goldman, Seitz, and Martin-Brualla}]{park2021nerfies}
Park, K.; Sinha, U.; Barron, J.~T.; Bouaziz, S.; Goldman, D.~B.; Seitz, S.~M.; and Martin-Brualla, R. 2021.
\newblock Nerfies: Deformable neural radiance fields.
\newblock In \emph{Proceedings of the IEEE/CVF International Conference on Computer Vision}, 5865--5874.

\bibitem[{Paszke et~al.(2019)Paszke, Gross, Massa, Lerer, Bradbury, Chanan, Killeen, Lin, Gimelshein, Antiga et~al.}]{paszke2019pytorch}
Paszke, A.; Gross, S.; Massa, F.; Lerer, A.; Bradbury, J.; Chanan, G.; Killeen, T.; Lin, Z.; Gimelshein, N.; Antiga, L.; et~al. 2019.
\newblock Pytorch: An imperative style, high-performance deep learning library.
\newblock \emph{Advances in neural information processing systems}, 32.

\bibitem[{Poole et~al.(2022)Poole, Jain, Barron, and Mildenhall}]{poole2022dreamfusion}
Poole, B.; Jain, A.; Barron, J.~T.; and Mildenhall, B. 2022.
\newblock DreamFusion: Text-to-3D using 2D Diffusion.
\newblock In \emph{The Eleventh International Conference on Learning Representations}.

\bibitem[{Roessle et~al.(2022)Roessle, Barron, Mildenhall, Srinivasan, and Nie{\ss}ner}]{roessle2022dense}
Roessle, B.; Barron, J.~T.; Mildenhall, B.; Srinivasan, P.~P.; and Nie{\ss}ner, M. 2022.
\newblock Dense depth priors for neural radiance fields from sparse input views.
\newblock In \emph{Proceedings of the IEEE/CVF Conference on Computer Vision and Pattern Recognition}, 12892--12901.

\bibitem[{Seo et~al.(2023)Seo, Han, Chang, and Kwak}]{seo2023mixnerf}
Seo, S.; Han, D.; Chang, Y.; and Kwak, N. 2023.
\newblock Mixnerf: Modeling a ray with mixture density for novel view synthesis from sparse inputs.
\newblock In \emph{Proceedings of the IEEE/CVF Conference on Computer Vision and Pattern Recognition}, 20659--20668.

\bibitem[{Shabanov et~al.(2024)Shabanov, Govindarajan, Reading, Goli, Rebain, Yi, and Tagliasacchi}]{shabanov2024banf}
Shabanov, A.; Govindarajan, S.; Reading, C.; Goli, L.; Rebain, D.; Yi, K.~M.; and Tagliasacchi, A. 2024.
\newblock BANF: Band-limited Neural Fields for Levels of Detail Reconstruction.
\newblock In \emph{Proceedings of the IEEE/CVF Conference on Computer Vision and Pattern Recognition}, 20571--20580.

\bibitem[{Shue et~al.(2023)Shue, Chan, Po, Ankner, Wu, and Wetzstein}]{shue20233d}
Shue, J.~R.; Chan, E.~R.; Po, R.; Ankner, Z.; Wu, J.; and Wetzstein, G. 2023.
\newblock 3d neural field generation using triplane diffusion.
\newblock In \emph{Proceedings of the IEEE/CVF Conference on Computer Vision and Pattern Recognition}, 20875--20886.

\bibitem[{Sun et~al.(2023)Sun, Zhang, Chen, Li, Ji, Zhao, and Xing}]{sun2023vgos}
Sun, J.; Zhang, Z.; Chen, J.; Li, G.; Ji, B.; Zhao, L.; and Xing, W. 2023.
\newblock VGOS: voxel grid optimization for view synthesis from sparse inputs.
\newblock In \emph{Proceedings of the Thirty-Second International Joint Conference on Artificial Intelligence}, 1414--1422.

\bibitem[{Trevithick and Yang(2021)}]{trevithick2021grf}
Trevithick, A.; and Yang, B. 2021.
\newblock Grf: Learning a general radiance field for 3d representation and rendering.
\newblock In \emph{Proceedings of the IEEE/CVF International Conference on Computer Vision}, 15182--15192.

\bibitem[{Truong et~al.(2023)Truong, Rakotosaona, Manhardt, and Tombari}]{truong2023sparf}
Truong, P.; Rakotosaona, M.-J.; Manhardt, F.; and Tombari, F. 2023.
\newblock Sparf: Neural radiance fields from sparse and noisy poses.
\newblock In \emph{Proceedings of the IEEE/CVF Conference on Computer Vision and Pattern Recognition}, 4190--4200.

\bibitem[{Verbin et~al.(2022)Verbin, Hedman, Mildenhall, Zickler, Barron, and Srinivasan}]{verbin2022ref}
Verbin, D.; Hedman, P.; Mildenhall, B.; Zickler, T.; Barron, J.~T.; and Srinivasan, P.~P. 2022.
\newblock Ref-nerf: Structured view-dependent appearance for neural radiance fields.
\newblock In \emph{2022 IEEE/CVF Conference on Computer Vision and Pattern Recognition (CVPR)}, 5481--5490. IEEE.

\bibitem[{Wang et~al.(2023)Wang, Chen, Loy, and Liu}]{wang2023sparsenerf}
Wang, G.; Chen, Z.; Loy, C.~C.; and Liu, Z. 2023.
\newblock Sparsenerf: Distilling depth ranking for few-shot novel view synthesis.
\newblock In \emph{Proceedings of the IEEE/CVF International Conference on Computer Vision}, 9065--9076.

\bibitem[{Wang et~al.(2021{\natexlab{a}})Wang, Liu, Liu, Theobalt, Komura, and Wang}]{wang2021neus}
Wang, P.; Liu, L.; Liu, Y.; Theobalt, C.; Komura, T.; and Wang, W. 2021{\natexlab{a}}.
\newblock NeuS: Learning Neural Implicit Surfaces by Volume Rendering for Multi-view Reconstruction.
\newblock \emph{Advances in Neural Information Processing Systems}, 34: 27171--27183.

\bibitem[{Wang et~al.(2021{\natexlab{b}})Wang, Wang, Genova, Srinivasan, Zhou, Barron, Martin-Brualla, Snavely, and Funkhouser}]{wang2021ibrnet}
Wang, Q.; Wang, Z.; Genova, K.; Srinivasan, P.~P.; Zhou, H.; Barron, J.~T.; Martin-Brualla, R.; Snavely, N.; and Funkhouser, T. 2021{\natexlab{b}}.
\newblock Ibrnet: Learning multi-view image-based rendering.
\newblock In \emph{Proceedings of the IEEE/CVF Conference on Computer Vision and Pattern Recognition}, 4690--4699.

\bibitem[{Wang, Skorokhodov, and Wonka(2023)}]{wang2023pet}
Wang, Y.; Skorokhodov, I.; and Wonka, P. 2023.
\newblock Pet-neus: Positional encoding tri-planes for neural surfaces.
\newblock In \emph{Proceedings of the IEEE/CVF Conference on Computer Vision and Pattern Recognition}, 12598--12607.

\bibitem[{Wei et~al.(2021)Wei, Liu, Rao, Zhao, Lu, and Zhou}]{wei2021nerfingmvs}
Wei, Y.; Liu, S.; Rao, Y.; Zhao, W.; Lu, J.; and Zhou, J. 2021.
\newblock Nerfingmvs: Guided optimization of neural radiance fields for indoor multi-view stereo.
\newblock In \emph{Proceedings of the IEEE/CVF International Conference on Computer Vision}, 5610--5619.

\bibitem[{Yang, Pavone, and Wang(2023)}]{yang2023freenerf}
Yang, J.; Pavone, M.; and Wang, Y. 2023.
\newblock Freenerf: Improving few-shot neural rendering with free frequency regularization.
\newblock In \emph{Proceedings of the IEEE/CVF Conference on Computer Vision and Pattern Recognition}, 8254--8263.

\bibitem[{Yu et~al.(2021)Yu, Ye, Tancik, and Kanazawa}]{yu2021pixelnerf}
Yu, A.; Ye, V.; Tancik, M.; and Kanazawa, A. 2021.
\newblock pixelnerf: Neural radiance fields from one or few images.
\newblock In \emph{Proceedings of the IEEE/CVF Conference on Computer Vision and Pattern Recognition}, 4578--4587.

\bibitem[{Zhao et~al.(2024)Zhao, Fan, Liao, and Yan}]{zhao2024grounding}
Zhao, Z.; Fan, F.; Liao, W.; and Yan, J. 2024.
\newblock Grounding and Enhancing Grid-based Models for Neural Fields.
\newblock In \emph{Proceedings of the IEEE/CVF Conference on Computer Vision and Pattern Recognition}, 19425--19435.

\bibitem[{Zhu et~al.(2024)Zhu, He, Li, Li, and Chen}]{zhu2024vanilla}
Zhu, H.; He, T.; Li, X.; Li, B.; and Chen, Z. 2024.
\newblock Is vanilla MLP in neural radiance field enough for few-shot view synthesis?
\newblock In \emph{Proceedings of the IEEE/CVF Conference on Computer Vision and Pattern Recognition}, 20288--20298.

\end{thebibliography}

\end{document}